\RequirePackage{fix-cm}
\documentclass{article}
\usepackage{graphicx}
\usepackage{mathptmx}      
\usepackage{graphicx}
\usepackage{caption}
\usepackage{subcaption}

\usepackage{algorithm}
\usepackage{algorithmic}
\usepackage{amsmath}
\usepackage{amssymb}
\usepackage{mathtools}
\usepackage{float}
\usepackage{hyperref}
\usepackage[natbibapa]{apacite}
\usepackage{comment}
\usepackage{a4wide}

\usepackage{etoolbox}
\makeatletter
\providecommand{\institute}[1]{
  \apptocmd{\@author}{\end{tabular}
    \par
    \begin{tabular}[t]{c}
    #1}{}{}
}
\makeatother

\begin{document}

\title{Exploring Offline Policy Evaluation for the Continuous-Armed Bandit Problem
}


\author{Jules Kruijswijk         \and
        Petri Parvinen \and
        Maurits Kaptein 
}

\date{}

\institute{Tilburg University \and Aalto University \and Tilburg University}

\maketitle

\begin{abstract}
The (contextual) multi-armed bandit problem (MAB) provides a formalization of sequential decision-making which has many applications. However, validly evaluating MAB policies is challenging; we either resort to simulations which inherently include debatable assumptions, or we resort to expensive field trials. Recently an \emph{offline} evaluation method has been suggested that is based on empirical data, thus relaxing the assumptions, and can be used to evaluate multiple competing policies in parallel. This method is however not directly suited for the continuous armed (CAB) problem; an often encountered version of the MAB problem in which the action set is continuous instead of discrete. We propose and evaluate an extension of the existing method such that it can be used to evaluate CAB policies. We empirically demonstrate that our method provides a relatively consistent ranking of policies. Furthermore, we detail how our method can be used to select policies in a real-life CAB problem. 
\end{abstract}

\section{Introduction}

In the canonical multi-armed bandit (MAB) problem a gambler stands in front of a row of slot machines, each with a (potentially) different payoff. It is up to the gambler to decide in sequence which machine to play and, during the course of sequentially playing the machines, she aims to make as much profit as possible by simultaneously learning from the previous observations and using the gained knowledge to steer future actions \citep{whittle1980,berry1985bandit}. 
The gambler needs to pick a \emph{strategy} that dictates which arm to play next given the previous observations. 

The problem of finding such a strategy is complicated since at each interaction the gambler only observes the outcomes of the machine she played, and she will never know the outcomes of the other possible courses of action at that moment in time. This so-called omission of \emph{counterfactuals} \citep{Li2011} -- not being able to gain knowledge about all the possible outcomes -- gives rise to the exploration versus exploitation trade-off \citep{berry1985bandit}: at each time point an action can either be geared at gaining more knowledge regarding the machines she is uncertain about (exploration), or it can be geared at using the knowledge gained in earlier interactions by playing machines with a high expected pay-off (exploitation). A good strategy balances this trade-off and does not waste too many plays on gaining new knowledge, nor does it become too greedy and get stuck exploiting a suboptimal machine \citep{kaelbling1996}.

The MAB problem is easily extended to more general settings. One such extension is the contextual MAB (cMAB) problem \citep{langford2008}. In the cMAB problem, at each interaction, the gambler observes the state of the world (context), which might influence the optimal choice at that moment in time \citep{langford2008,Beygelzimer2011,bubeck2012regret}. Both the MAB problem and the cMAB problem have been heavily analyzed \citep{wang2005bandit}. Furthermore, strategies, or in this literature more often called \emph{policies}, to address the (c)MAB problem have found many practical applications in recent years: examples include, but are not limited to, the personalization of online news \citep{Li2010b}, online advertisement selection \citep{cheng2010}, website morphing \citep{Hauser2009}, and adaptive clinical trials \citep{Press2009,williamson2017bayesian}. 

In the current article we focus on another extension of the MAB formalization coined the continuous-armed bandit (CAB) problem \citep{Agrawal1995}: this problem distinguishes itself apart from other formalizations by considering instead of a set of discrete actions (the distinct slot machines) a continuous range of actions. The CAB problem has also been analyzed \citep{Agrawal1995,krause2011contextual,kleinberg2004nearly}, however the current theory focussed literature lacks applied methods to evaluate the performance of different CAB policies. This is true despite many applied settings in which this problem is encountered. Which include but are not limited to, choosing an optimal price for selling a product to customers encountered sequentially, or choosing an optimal treatment dose. In this paper we suggest and evaluate a practical method for evaluating the performance of different CAB strategies in an externally valid setting. 

As stated above, the omission of counterfactuals complicates finding a good policy since it gives rise to the exploration-exploitation trade-off. Counterfactuals also complicate the evaluation of competing policies: due to the omission of counterfactuals in the collected data resulting from field evaluations of a specific policy, it is challenging to use these existing data to directly evaluate alternative strategies. Therefore, if we want to empirically evaluate bandit policies we either have to resort to running multiple field evaluations that are often very expensive to carry out, or we have to resort to simulation based methods which often lack external validity. A few years ago, \citet{Li2011} suggested an effective solution to this problem for the (c)MAB problem: they proposed a method for the externally valid \emph{offline} -- thus based on existing, pre-collected, data -- evaluation of MAB policies. The method relies on a single dataset, thus cutting costs, while it circumvents the validity problems that easily arise in simulations by using actual empirical data. 

The offline MAB evaluation method suggested by \citet{Li2011} relies on collecting -- in the field -- a dataset in which the actions where taken uniformly at random at each interaction. Next, to evaluate a particular decision policy, the sequence of data points is replayed and, at each interaction, the action suggested by the policy under evaluation is compared to the action that is actually present in the logged data at that point in the sequence. If the two actions match, the data point gets ``accepted'' and its outcome is included in the evaluation of the policy. If the actions do not match the interaction is simply ignored. This method demonstrably provides unbiased estimates of the performance of distinct bandit policies (albeit for a smaller number of interactions than the number of datapoints collected in the initial field trial).

In practice, the method by \citet{Li2011} works well only when the number of actions is (relatively) low, the amount of observations is large, or both. If however the action set is large, or the number of observations in the dataset small, estimates of the performance of the different policies can only be obtained for a (very) small number of interactions. Cleary, in the limit, the suggested method thus fails for the CAB problem; since theoretically the number of possible actions is infinite, the probability that the actions suggested by a policy under evaluation matches the randomly selected action in the existing dataset tends to zero. As a result, no observations will be accepted and the evaluation of the policy fails. 

Other methods for offline evaluation exist such as the method introduced by \citet{thomas2015high}. The advantage of this method is that the dataset does not have to be collected uniformly randomly, as long as the probability of the action that was played is known. This works well in the MAB setting. In the CAB setting however, calculating the this probability is not trivial - since calculating the probability of playing a certain action tends to zero, as described above. This means that we cannot readily use this method for the evaluation of policies for the CAB problem as well.

In this paper, we present and empirically evaluate, on a practical scale, a logical extension to the method of \citet{Li2011} to make it suitable for the evaluation of CAB policies. In the next section, we first detail the method developed by \citet{Li2011}. Next, we introduce our extension to the method and discuss its rationale. Then, we empirically evaluate the performance of the suggested method by showing that it allows one to consistently order bandit policies for CAB problems for multiple sizes of the problem and for multiple true data generating models. While our suggested method does not provide unbiased estimates of the absolute performance of bandit strategies -- we explicate this in the next section -- it does provide a cheap and straightforward method to provide a relative rank of distinct policies and thus aid decision making when selecting policies for applied CAB problems. Finally, we present the use of our method in the field of online marketing and discuss future research directions and possible improvements of our method. 

\section{Offline CAB policy evaluation}

Before we introduce our extension to the method proposed by \citet{Li2011}, we first more formally introduce the (c)MAB (and CAB) problem. 
Bandit problems can be described as follows: at each time $t = 1,...,T$, we have a set of possible actions $\mathcal{A}$. After choosing $a_t \in \mathcal{A}$ we observe reward $r_t$. The aim is to find a \emph{policy} ($\Pi(h_{t-1})$ where $h_{t-1}$ is the historical data), which is a mapping from all the historical data to the action at $t$, to select actions such that the cumulative reward $R_c =  \sum_{t=1}^T r_t$ is as large as possible.\footnote{Note that a slightly more general formulation of the problem does not consider a fixed horizon T but rather some discount rate for future rewards \citep{berry1985bandit}.}  In the case of a CAB problem, the same formalization can be used, where the only difference is in the action set: in the CAB problem we have $\mathcal{A} \in \mathbb{R}$ (often constrained within some range $[a,b]$). 

To assess how a policy performs we often look at the expected regret of the policy which is defined by
\begin{equation}
\mathbb{E}(R_{T}) = \sum_{t=1}^{T} r^{*}_t - r_t
\label{eq:regret}
\end{equation}
where $r_t$ is the reward at interaction $t$ and $r^{*}_t$ is the reward of the action with the highest expected
reward (the optimal policy). Regret, as opposed to the cumulative reward, $R_c$, provides an intuitive benchmark since a perfect strategy would incur an expected regret of $0$. Further note that if a suboptimal action has a non-zero and non-decreasing probability of being selected, the regret will -- in expectation -- increase linearly. Most analytical work focusses on showing the asymptotic sub-linear regret of distinct (c)MAB or CAB policies \citep[e.g.,][]{Lai1985}. 

In many applied situations, however, we have no knowledge about the actions with the highest expected reward (i.e. we do not know $r^{*}$) and thus we will not be able to compute the regret. In such cases the best we can do is compare the cumulative reward $R_c$ (or the average per time point reward $R_c / T$) obtained over multiple comparable runs -- either in simulations or in field evaluations -- of the policies under evaluation. However, this highlights a clear challenge when evaluating multiple policies: simulations likely contain assumptions that limit the external validity of the evaluation, while in-field evaluations of multiple policies are often difficult and expensive to carry out. 

To address these problems \citet{Li2011} proposed a method to obtain unbiased estimates of the expected cumulative reward of different policies using a single, externally valid, dataset. Algorithm \ref{alg:li_eval} details the proposed method: we run sequentially through a stream of logged data in which the actions have been selected uniformly at random. At each event in the stream, the policy under evaluation proposes an action. If the action proposed by the policy is the same as the action of the logged event, then the event is counted towards the evaluation of the policy and the observed reward is added to the total payoff. Note that if there are $K$ actions, then the number of valid events $T$ in the evaluation process is a random number with expected value $L/K$, where $L$ is the length of the logged data set. Thus, during the evaluation datapoints are ``accepted'' with probability $p_{accept} = \frac{1}{K}$. In the online setting, $L = T$ and in the offline setting $\mathbb{E}(T) = \frac{L}{K}$.

\citet{Li2011} show that, under a number of assumptions regarding the collection of the logged dataset and the stationarity of the process, the method described in Algorithm \ref{alg:li_eval} provides an unbiased estimate of the performance of policy $\Pi$. As such, the method makes it possible to compare multiple competing policies in an externally valid setting without the recurring costs of repeating field trials. In practice, however, the method fails for the often encountered CAB problem. This is due to the fact that with continuous action space the probability that a logged action is equal to a suggested action by the policy is very low: as $K$ grows $p_{accept}$ decreases and we have for the CAB problem $K \rightarrow \infty$ and $p_{accept} \rightarrow 0$. Thus, the method fails. 

\begin{algorithm}[H]
    \centering
\caption{Policy evaluator with finite data stream for the MAB problem.}
\label{alg:li_eval}
\begin{algorithmic}
   \large
    \STATE Inputs: policy $\Pi$; stream of events $S$ of length $L$
    \STATE $h_0 \leftarrow \empty$ (An initially empty history)
    \STATE $R_c \leftarrow 0$ (An initially zero total payoff)
    \STATE $T \leftarrow 0$ (An initially zero counter of valid events)
    \FOR{$t = 1,2,\dots,L$}
    	\STATE Get the $t$-th event $(a, r_a)$ from $S$
	\IF{$\Pi(h_{t-1}) = a$}
	    \STATE $\text{update } \Pi(r_{a},a)$
	    \STATE $h_t \leftarrow \text{CONCATENATE}(h_{t-1},(a,r_a))$
	    \STATE $R_c \leftarrow R_c + r_a$
	    \STATE $T \leftarrow T + 1$
	\ELSE
	    \STATE $h_{t} \leftarrow h_{t-1}$
	\ENDIF
    \ENDFOR
    \STATE Output: $R_c$ and $T$ (or $R_c/T$ for the average reward)
\end{algorithmic}
\end{algorithm}

In an attempt to solve this problem and to provide a practically usable method for the offline evaluation of CAB policies we propose an alternative to the method suggested by \citet{Li2011}. Algorithm \ref{alg:cab_eval} describes our logical adaptation of Algorithm \ref{alg:li_eval} to provide an evaluation method for the CAB problem. The difference between the two algorithms is in the \textbf{if} statement that determines acceptance of the proposed action: instead of constraining the
suggested action to be \emph{exactly} equal to the logged action, we compare the distance between the action logged in the dataset, $a$, with the action proposed by the policy, $\Pi(h_{t-1})$. If the absolute distance between these two actions is less than the tuning parameter $\delta$, we accept the data point, and else it will be discarded. Intuitively the proposed change corresponds to the difference between the evaluation of a PDF of a discrete versus a continuous random variable. Note here that we update the policy not with the logged action $a$ but rather with the action that was proposed by the policy ($\Pi(h_{t-1})$). Hence providing a noisy estimate of the reward at $\Pi(h_{t-1})$.

\begin{algorithm}[H]
     \centering
\caption{Offline policy evaluation for the CAB problem}\label{euclid}
\label{alg:cab_eval}
\begin{algorithmic}
        \large
	\STATE Inputs: policy $\Pi$; stream of events $S$ of length $L$ with actions selected randomly in the range $[a,b]$
	\STATE $h_0 \leftarrow \empty$ (An initially empty history)
	\STATE $R_c \leftarrow 0$ (An initially zero total payoff)
	\STATE $T \leftarrow 0$ (An initially zero counter of valid events)
	\FOR{$t = 1,2,\dots,L$}
	        \STATE Get the $t$-th event $(a, r_a)$ from $S$
		\IF{$|a - \Pi(h_{t-1},)| < \delta$}
		\STATE $\text{update }\Pi(r_a,\Pi(h_{t-1}))$  
	        \STATE $h_t \leftarrow \text{CONCATENATE}(h_{t-1},(\Pi(h_{t-1}),r_a))$
	        \STATE $R_c \leftarrow R_c + r_a$
	        \STATE $T \leftarrow T + 1$
		\ELSE
		\STATE $h_{t} \leftarrow h_{t-1}$
		\ENDIF
	\ENDFOR
	 \STATE Output: $R_c$ and $T$
\end{algorithmic}
\end{algorithm}

\subsection{Properties of the offline CAB evaluation}

Before we empirically evaluate the applied use of proposed method in an extensive simulation study in the next section, it is worthwhile to analyze the role of the tuning parameter $\delta$ and to reflect on the resulting estimates of $R_c$ that follow from our procedure. This is most easily done by keeping in mind a very simple CAB formalization where the true data generating process is merely a parabola constrained within the range $[0,1]$, say $r_t = f(a_t) = -(a_t - .5)^2 + \epsilon$ where $\epsilon$ represents some random noise and we have $E(\epsilon) = 0$. Note that $\Pi^{*}(t) = a^*_t = .5$. Clearly, a large value of $\delta$ (e.g., $.25$) will lead to accepting a high number of proposed actions (and thus large number of evaluations $T$), but will also lead to high variation in the realizations of $f(a_t)$: the policy evaluates $f()$ at $\Pi(h_{t-1},x)$, and receives as a result $f(a_t)$ which might be at most $\delta$ away. Hence, for large $\delta$, the performance of the policy will be poor since it obtains erroneous evaluations of $f(a_t)$, and the estimated cumulative regret will be (severely) biased. The exact way in which a policy will be biased by these erroneous evaluations of $f()$ heavily depends on the way in which the policy incorporates the history $h_{t-1}$ when selecting the next action. These problems diminish as $\delta$ decreases, however, the number of accepted observations $T$ will decrease accordingly. Hence, $\delta$ should be chosen as small as possible, however the expected number of accepted events $T = p_{accept} L = \frac{2\delta}{b-a} L$ (with range of actions $[a,b]$) should be as close as possible to the expected number of events occurred in the real-life setting for which the policy is evaluated. This implies that in practice one would like to collect a dataset containing uniformly randomly selected actions in range $[a,b]$ with length $L = \frac{2\delta}{(a-b)/T'}$ where $T'$ is the desired length of the policy evaluation for the applied problem. 

Note that as long as $\delta > 0$ the estimated expected reward $E(R_c)$ is downwardly biased (for concave functions) since even if the policy converges exactly on the optimal action (e.g., selecting $a_{t=t',\dots,t=T} = a^* = .5$ for some $t'$) the evaluations of $f(a_t)$ originate uniformly randomly from the interval $[a^*-\delta, a^*+\delta]$. Since each evaluation for which $a_t \neq a^*$ leads, in expectation, to a reward $r_t \leq r^*_t$, the expected cumulative reward of the policy under consideration is downwardly biased. Nonetheless, for the comparison of the \emph{relative} performance of applied CAB policies this is not as cumbersome as it might sound; as long as the ranking of policies is relatively consistent for the desired scale of the problem $T$ the method is still useful to select one out of a number of competing policies. We will demonstrate below that this is the case for a range of values of $\delta$ as well as for multiple true data generating functions $f()$. Hence, our proposed method is valuable in practice. 

\section{Evaluating offline CAB policy evaluation using simulation}

To evaluate our proposed method we perform 2 simulation studies in which, using different true underlying models $f()$, we first examine the performance of a set of $4$ CAB policies while collecting data ``online'' (e.g., when simulating the rewards directly from our known $f()$). Next, we collect an offline evaluation dataset $S$ by gathering rewards $r_{1,\dots,L}$ from $f()$ for actions choosing uniformly at random: $a_{1\dots,L} \sim \text{Unif}[a,b]$. Subsequently, with different choices of $\delta$, we evaluate the same $4$ CAB policies using our proposed offline procedure. Finally, we compare the performance in terms of expected cumulative regret over multiple simulations runs, between the online and offline procedures. 

\subsection{Shared Methods for the two simulation studies}

To carry out the simulation experiment we used \texttt{StreamingBandit} (see \url{https://github.com/Nth-iteration-labs/streamingbandit}). StreamingBandit is a python web application aimed at developing and testing bandit policies in field studies. It is very flexible, as it allows for alternating, nesting and comparing different
policies in real time \citep{Kaptein16}. 

The $4$ policies examined in our simulation studies to evaluate our proposed method for the offline evaluation of CAB policies are:
\begin{enumerate}
	\item The \emph{Uniform random, UR,} policy. In this policy simply $a_{1,\dots,T} \sim \text{Unif}(a,b)$. The regret of this policy is expected to grow linearly.
	\item The \emph{$\epsilon$-first, EF,} policy. This is a greedy algorithm that has two phases. In the first
	phase, which is restricted by a preset $N$ number of interactions, the
	policy \emph{explores}: $a_{1,\dots,N} \sim \text{Unif}(a,b)$. Next, a simple linear model is fit to the observed data. The model that we fit using standard least squares estimation is
	\begin{eqnarray}
	r = \beta_0 + a\hat{\beta}_1 + a^2  \hat{\beta}_2 
	\label{eq:model}
	\end{eqnarray}
	and subsequently, in the exploit stage we choose the action that maximizes this fitted curve, $a_{N+1,\dots,T} = \frac{-\hat{\beta}_1}{2\hat{\beta}_2}$. The regret of this policy is expected to grow linearly in both phases, however, in the first phase it will grow faster than in the second phase, since it uses the expected (heuristically) optimal action in the second phase (and stops exploring).
	\item The \emph{Thompson sampling using Bayesian linear regression, TBL,} policy. Thompson sampling \citep{Thompson1933, Scott2010, Agrawal2011} is a sampling method in which an action $a_t$ is randomly selected with a probability proportional to the belief that this action is the best action to play given some (Bayesian) model of the relationship between the actions and the rewards. Consider the all the historical data, previously denoted $h_t$, consisting of the history of the actions and rewards up to $t$. Further denote the parameters $\theta = \{\beta_0, \beta_1, \beta_2\}$ (as in Eq. \ref{eq:model}). We setup a Bayesian model using some prior on $P(\theta)$ and obtain posterior $P(\theta | h_t) \propto P(h_t | \theta)P(\theta)$. To subsequently select an action proportional to its probability of being optimal -- and thus to implement Thompson sampling -- it suffices to obtain a single draw $\theta'_t$ from the posterior $P(\theta | h_t)$ and then select the action that is optimal  given the current draw using $a_t = \frac{-\hat{\beta'}_1}{2\hat{\beta'}_2}$ \citep{Scott2010}. To compute $P(\theta | h_t)$ we use -- using matrix notation --  the well known online Bayesian linear regression model \citep[as described, e.g., by][]{box2011bayesian} where we update at each time point:
	        \begin{eqnarray}
	        J & := &  J + \frac{r_t \boldsymbol{a}^{T}}{\sigma^2},\\
		P & := & P + \frac{\boldsymbol{a}\boldsymbol{a}^{T}}{\sigma^2}
		\end{eqnarray}
		and where $J = \Sigma^{-1} \mu$, $P = \Sigma^{-1}$ and $\boldsymbol{a} = [1, a, a^2]$. Finally, the sought after
		${\theta | h_t} \sim \mathcal{N}(\mu, \Sigma)$ from which draws are easily obtained at each time point. We again use the model presented in Eq. \ref{eq:model} and we use the starting values $J =  [0,0.05,-0.05]$ and $P = \text{diag}(2,2,5)$. The regret of this policy is expected to grow sub-linearly.
	\item The \emph{Lock-in Feedback, LiF,} policy. LiF is a novel algorithm developed by
		\citet{Kaptein2015} \citep[see also][]{kaptein2016tracking}. LiF is inspired by a method that is
		frequently used in physics, coined lock in amplification \citep{wolfson1991lock,Scofield1994} that is routinely used to find -- and lock in to -- optima of noisy signals. LiF works by oscillating sampled values with a known frequency and amplitude around an initial value $a_0$. Using the observed feedback from the oscillations in the evaluations of $f()$ it is straightforward to find the derivative $f'()$ at $a_0$ and use a gradient ascent updating scheme to find $a^*_t$, see Algorithm \ref{al:lif} for details. During the simulations we choose starting values $a_0 \sim \mathcal{U}(0,1)$, $A = 0.05$, $i = 50$, $\gamma = .1$, $\omega = 1$. This function is expected to grow linearly in terms of regret, although it is expected that it grows slower than the first two policies (i.e., it reaches an optimum faster, but because of its oscillating nature the expected regret keeps growing linearly).
\end{enumerate}   

\begin{algorithm}
\caption{Implementation of the LiF policy as used in our evaluations. Here $T$ denotes the total length of the data stream of ``accepted'' actions.}
\label{al:lif}
\begin{algorithmic}
\large
\STATE Inputs: value $a_0$, amplitude $A$, integration window $i$, learn rate $\gamma$, and frequency $\omega$
\STATE $r_{\omega}^{\Sigma} \leftarrow 0$ (cumulative rewards)			  
\FOR{$t=1, \dots, T$}
	\STATE $a_t = a_0 + A \cos{\omega t}$									         
	\STATE $r_t = f(a_0 + A \cos{\omega t}) + \epsilon_t$						          
	\STATE $r_{\omega}^{\Sigma} = r_{\omega}^{\Sigma} + r_t \cos{\omega t}$	    
	\IF{$(t \mod i == 0)$}
		\STATE $r_{\omega}^{*} =  r_{\omega}^{\Sigma} / i$					    
		\STATE $a_0 = a_0 + \gamma r_{\omega}^{*} 	$				
		\STATE $r_{\omega}^{\Sigma} = 0$						 
	\ENDIF
\ENDFOR
\end{algorithmic}
\end{algorithm}

The above policies where chosen to a) include a very naive benchmark (the UR policy), and a number of different approaches advocated in the (c)MAB or CAB literature \citep{Sutton,Kaptein2015,box2011bayesian}. Please note that the number of possible alternative policies we could have explored is extremely large, ranging from simple heuristic strategies such as $\epsilon$-greedy \citep{Sutton} to currently popular Gaussian processes \citep{djolonga2013high}; we hope however to have included a selection of policies that provides an informative evaluation of the merits of our proposed method.  

\begin{figure}[htb]
\centering
\begin{subfigure}[b]{0.5\textwidth}
	\includegraphics[width=\textwidth]{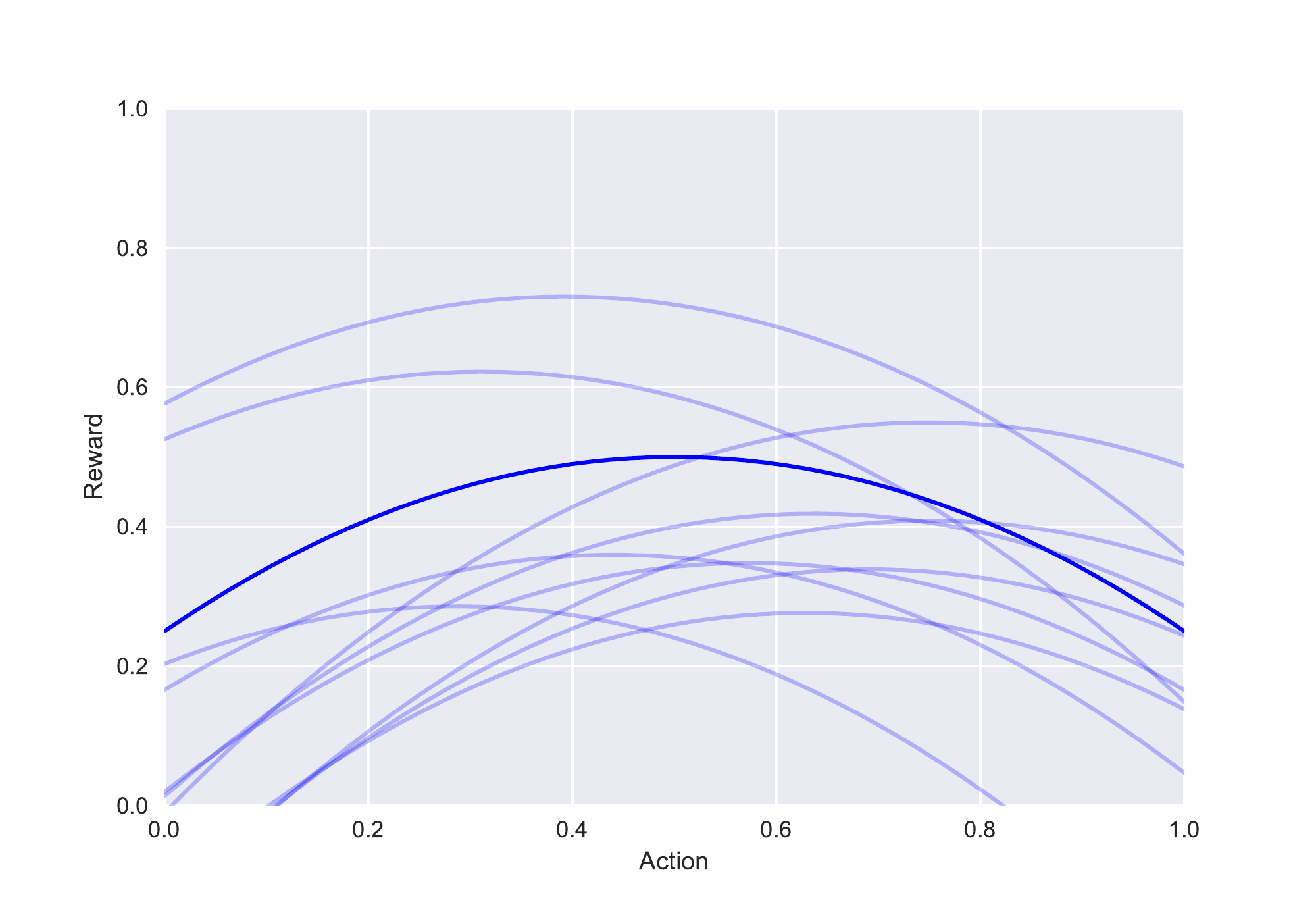}
\end{subfigure}%
\begin{subfigure}[b]{0.5\textwidth}
	\includegraphics[width=\textwidth]{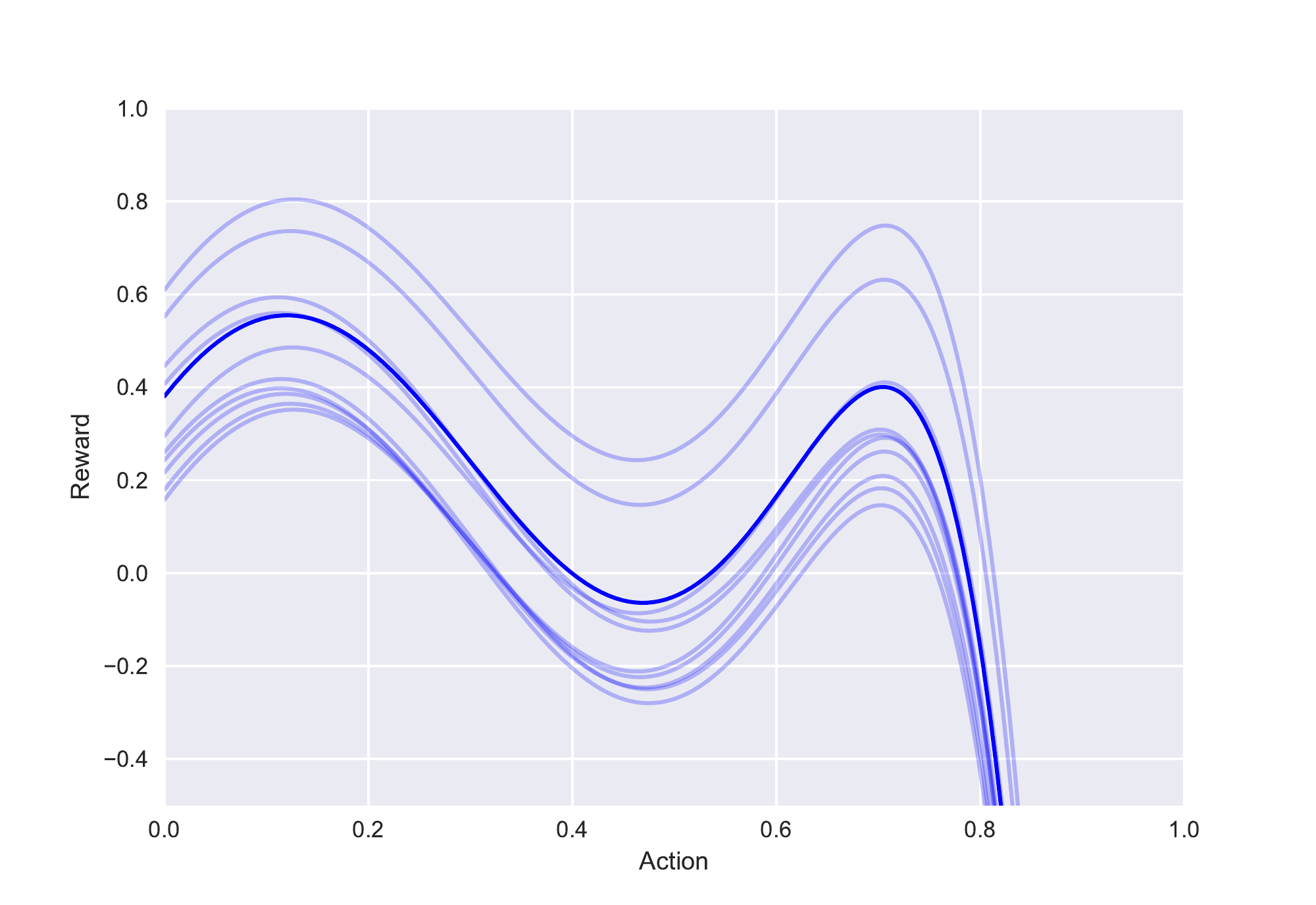}
\end{subfigure}
	\label{fig:complot}
\caption{The true reward functions used in study 1 (left) and study 2 (right).}
\label{fig:rewards}
\end{figure}

The two simulation studies differed in their specification of the true reward function $f()$. The two reward functions used were both varied across simulation runs to not favor particularities of the distinct policies under scrutiny. The first true reward function, $f_1()$, was a simple parabola defined between $[0,1$] with a maximum that changed randomly each simulation run. This scenario was chosen to represent simple one-dimensional CAB problems where a single maximum exists. The second simulation study was performed using a randomly generated higher order polynomial, $f_2()$, that results in a bimodal function in the range $[0.1]$. This function was chosen to represent a more complex single dimensional scenario where multiple maxima can be expected. The exact location of the two maxima, and the difference in height between the two maxima, differed for each simulation run. Figure \ref{fig:rewards} presents several instances of the reward functions used in the first and the second study respectively. Note that in the Figure the reward functions are presented without noise; in the simulations reward where given by $r = f() + \epsilon$ where $\epsilon \sim \mathcal{N}(0, 0.01)$.
 
In each simulation study we first ran $1,000$ repetitions of $L=T=10,000$ interactions to obtain online estimates of the performance of the $4$ policies; this provides our benchmark. Next, we ran $1,000$ repetitions, each on a data set $S$ of length $L=10,000$ with different values of $\delta \in (.01, .05, .1, .2, .5)$, of Algorithm \ref{alg:cab_eval} for each of the $4$ policies under consideration. This results in multiple offline evaluations of the same policies which we subsequently compare with the true online performance. Note that the total length of the offline evaluations differs for different values of $\delta$. Also note that the reward functions $f()$ also differed for the online and offline situations, each repetition we randomly generated a different $f()$ (as explained above). In the simulation studies we have chosen for a smaller practical scale in terms of number of interactions, as a simulation study on a larger scale might be interesting to investigate the properties for an infinite horizon for $T$, but a smaller scale is more useful for situations where we know that the amount of data will be rather small.

\subsection{Results of Simulation Study 1: The simple model}

\begin{figure}[htb]
\centering
	\includegraphics[width=\linewidth]{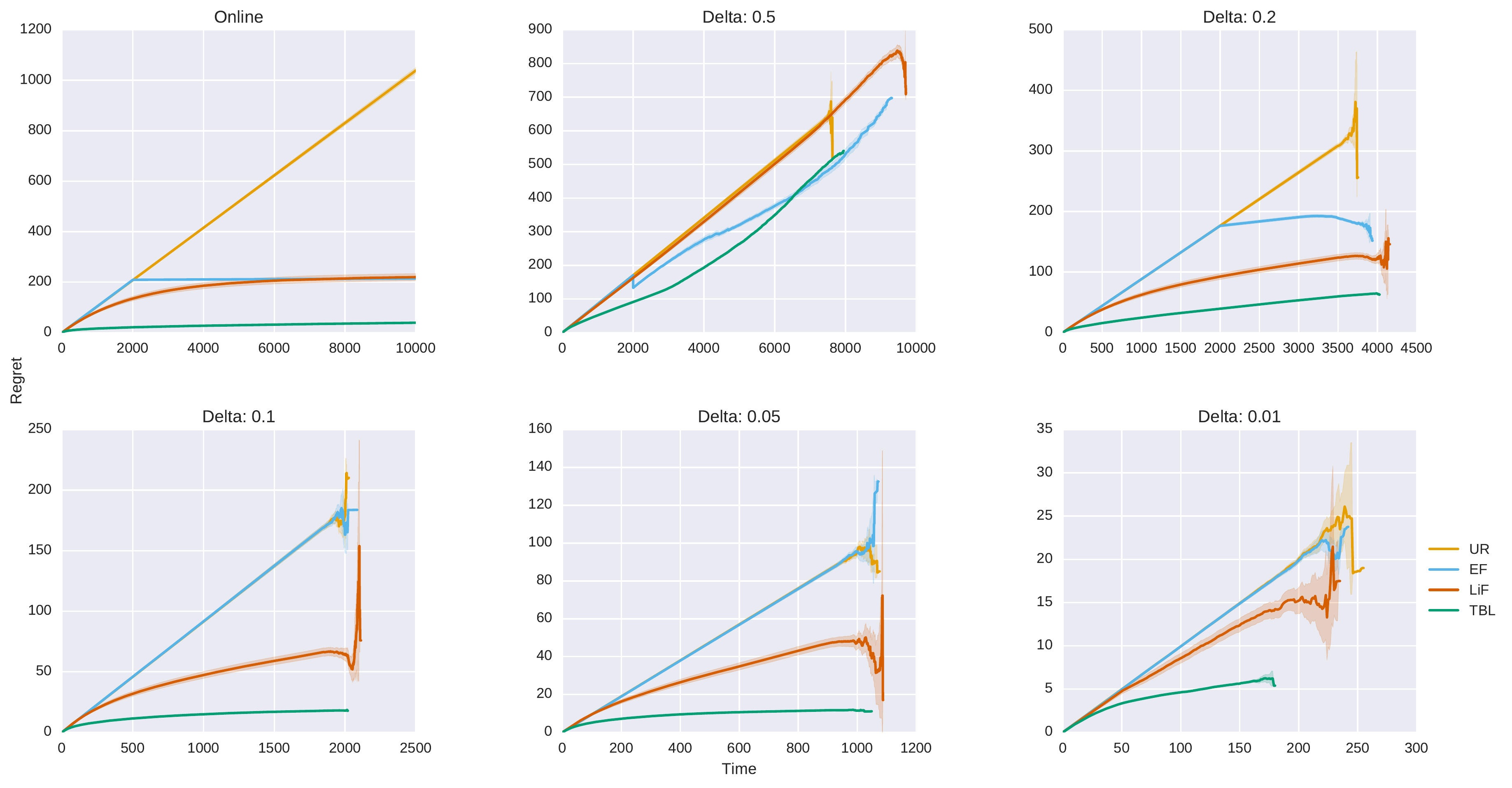}
\caption{The results for simulation Study 1 using a simple unimodal data generating model in terms of cumulative regret averaged over the simulation runs. The upper left panel shows the performance of the four policies for the online simulation. The other panels show the performance for the four policies for the offline evaluation with varying $\delta$'s. The lines are plotted together with their $95\%$ confidence bounds. Duly note that the confidence bounds can be misleading, since for low $\delta$'s we have a low amount of $T$ and also only a few repetitions left to average over (as compared to higher $\delta$'s).}
\label{fig:simple}
\end{figure}

Figure \ref{fig:simple} shows the results for the simple model in terms of empirical regret (see Equation \ref{eq:regret}). In the upper left panel, shows the performance of the four policies in the online simulation studie. Here it is clear that TBL performs very well and evaluates the function close to its maximum relatively quickly leading to a small regret. Also LiF seems to converge, but, due to its continuous oscillations, the regret keeps increasing; a linear increase is expected for this policy. EF performs as expected; first it incurs high regret due to the exploration stage, after which regret is small but linear in the exploitation stage. Given the relatively simple true reward function the exploitation stage often evaluates the reward function close to its maximum. The subsequent panels (left to right) show the performance of the distinct policies in terms of regret averaged over the $1,000$ simulation runs for decreasing values of $\delta$. Depicted are both the average regret over the $1,000$ simulation runs, as well as their empirical standard errors (the confidence bands). Note that the standard errors for the offline evaluations increase heavily towards the higher values of the plot; this is caused by the fact that higher values of $T$ become less and less likely in the offline evaluation. 

As expected, for large values of $\delta$ a large number of observations is obtained (e.g., $T$ is large), but the performance of the policies is severely affected by the extremely noisy evaluation of the true reward function. At $\delta = .5$ this results in an evaluation that provides hardly any information regarding the relative performance of the different policies. However, as $\delta$ decreases (and subsequently $T$ decreases), we find a more and more clear ordering of the policies. To illustrate, Table \ref{tab:rankorder} presents the relative rank order of the policies in terms of lowest regret evaluated at $T=1750$. The Table makes clear that our proposed offline CAB policy evaluation method consistently ranks the policies that are being compared.\footnote{Note that for $\delta=.5$ there is a small drop in regret for EF at $2,000$ observations. This is the time point at which EF moves from exploration to exploitation, and we find that in for a number of models the optimal action value is predicted outside the $[0-\delta,1+\delta]$ range and thus not included in the remaining runs.}

\begin{table}[htb]
\caption{Table displaying the rank order of the four policies under scrutiny at $T=1750$ for the online and offline evaluations for the simple model. Note that at this point by design UF and EF are in a tie. Further note that for small $\delta$, $T=1750$ is not observed.}
\label{tab:rankorder}
\begin{tabular}{lcccccc}
\hline\noalign{\smallskip}
Rank & Online & $\delta = .5$ & $\delta = .2$ & $\delta = .1$ & $\delta = .05$ & $\delta = .01$ \\
\noalign{\smallskip}\hline\noalign{\smallskip}
1    &   TBL     		&         TBL?      &      TBL         &     TBL          &        n/a        &       n/a         \\
2    &   LiF     		&         LiF?      &        LiF       &        LiF       &       n/a         &       n/a         \\
3    &   EF/UR     	&    EF/UR           &     EF/UR          &        EF/UR          &        n/a        &      n/a          \\
4    &   EF/UR     	&       EF/UR        &       EF/UR        &        EF/UR          &       n/a         &        n/a       \\
\hline\noalign{\smallskip}
\end{tabular}
\end{table}

\subsection{Results of Simulation Study 2: The complex model}

\begin{figure}[htb]
\centering
	\includegraphics[width=\textwidth]{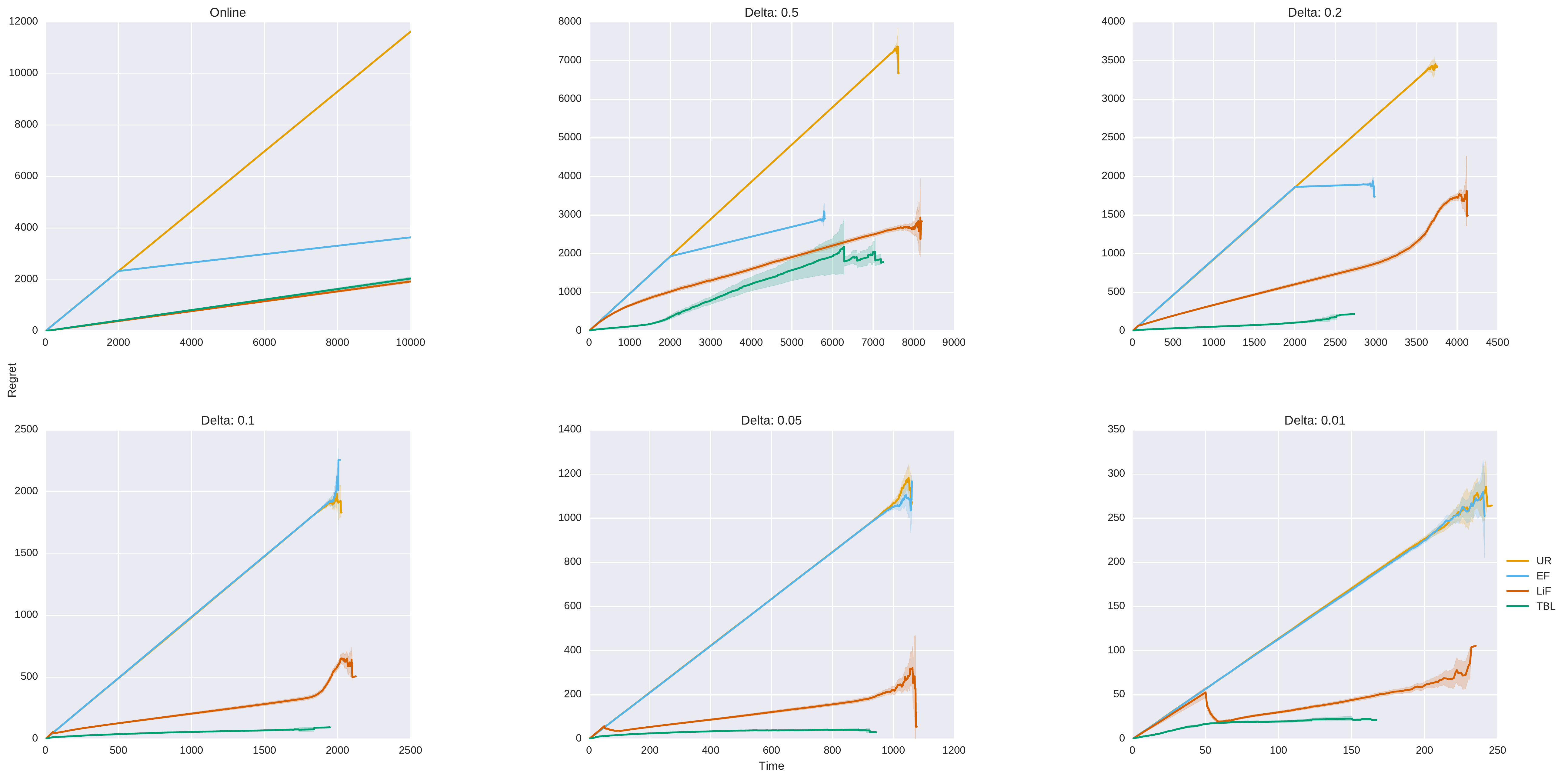}
\caption{The results for simulation Study 2 using a more complex data generating model in terms of cumulative regret. The upper left panel shows the performance of the four policies for the online simulation. The other panels show the performance for the four policies for the offline evaluation with varying $\delta$'s. The lines are plotted together with their $95\%$ confidence bounds. Duly note that the confidence bounds can be misleading, since for low $\delta$'s we have a low amount of $T$ and also only a few repetitions left to average over (as compared to higher $\delta$'s).}
\label{fig:complex}
\end{figure}

Figure \ref{fig:complex} shows the results for the more complex, bimodal, model. The first panel again shows the performance of the different policies in an online simulation. The Figure displays a similar pattern for the UR and EF policies as before: steep linear regret for the UR policy, and the same regret in the exploration stage for EF, after which in the exploitation stage the regret is lower; note that in this more complex case the optimum is not clearly found in the exploitation stages. Our implementation of TBL and LiF seem to have a comparable performance; note that given our current specification of the model used for both TBL (Eq. \ref{eq:model}) as well as the implementation of LiF both policies are likely to get ``stuck'' in a local maximum thus incurring returning (linear) regret.  

\begin{table}[htb]
\caption{Table displaying the rank order of the four policies under scrutiny at $T=1750$ for the online and offline evaluations for the complex model. }
\label{tab:rankorder2}
\begin{tabular}{lcccccc}
\hline\noalign{\smallskip}
Rank & Online & $\delta = .5$ & $\delta = .2$ & $\delta = .1$ & $\delta = .05$ & $\delta = .01$ \\
\noalign{\smallskip}\hline\noalign{\smallskip}
1    &   LiF      		&         TBL      &      TBL         &     TBL          &        n/a        &       n/a         \\
2    &   TBL     		&         LiF      &        LiF       &        LiF       &       n/a         &       n/a         \\
3    &   EF/UR     	&    EF/UR           &     EF/UR          &        EF/UR          &        n/a        &      n/a          \\
4    &   EF/UR     	&       EF/UR        &       EF/UR        &        EF/UR          &       n/a         &        n/a       \\
\hline\noalign{\smallskip}
\end{tabular}
\end{table}

Table \ref{tab:rankorder2} again displays the relative rank ordering of the policies. Note that in this case again a clear -- and correct -- separation is visible between the EF and UR policies and the TBL and LiF policy: TBL and LiF are clearly preferred. However, TBL and LiF seem competitive in the online evaluation, while offline clearly TBL is preferred; apparently TBL is more robust to the noise introduced by the offline evaluation method.\footnote{Note a similar drop-off for LiF as we saw with EF in the simple model at lower $\delta$'s; again, the policy ventures outside the selected range.} In any case however, the offline evaluation would lead one to select a policy that performs relatively well on the current problem. 

\section{Evaluation of our method in the field}

After evaluating the performance of our method, we demonstrate how it can be used in practice. In collaboration with a company that (re-)sells products online by offering rebates for online stores we collected a dataset for offline CAB policy evaluation. The company negotiates deals with existing online stores, and offers a share of these deals to its customers. For example, an online store that sells sportswear can offer a $10\%$ total discount to the  rebate company. Next, the rebate company splits this discount between herself and the end-customer. The current practice is to split the discount 50-50: if a customer wants to buy a pair of shoes from a online store that cost $50\$$, the rebate company receives a total $5\$$ cashback from the online store and gives $2.5\$$ to the customer. 

The company, however, does not know whether the 50-50 split actually maximizes their revenue. This gives rise to a CAB problem in which the action consists of the rebate percentage offered and the rewards consists of the revenues generated. Note that it is expected that a single maximum of this ``split-revenue'' function exists: passing a large part of the discount on to the customer likely leads to a large volume, but little revenue for the rebate company, while passing on a small discount likely leads to a smaller volume. 

\begin{figure}[htb]
\centering
\includegraphics[width=0.75\textwidth]{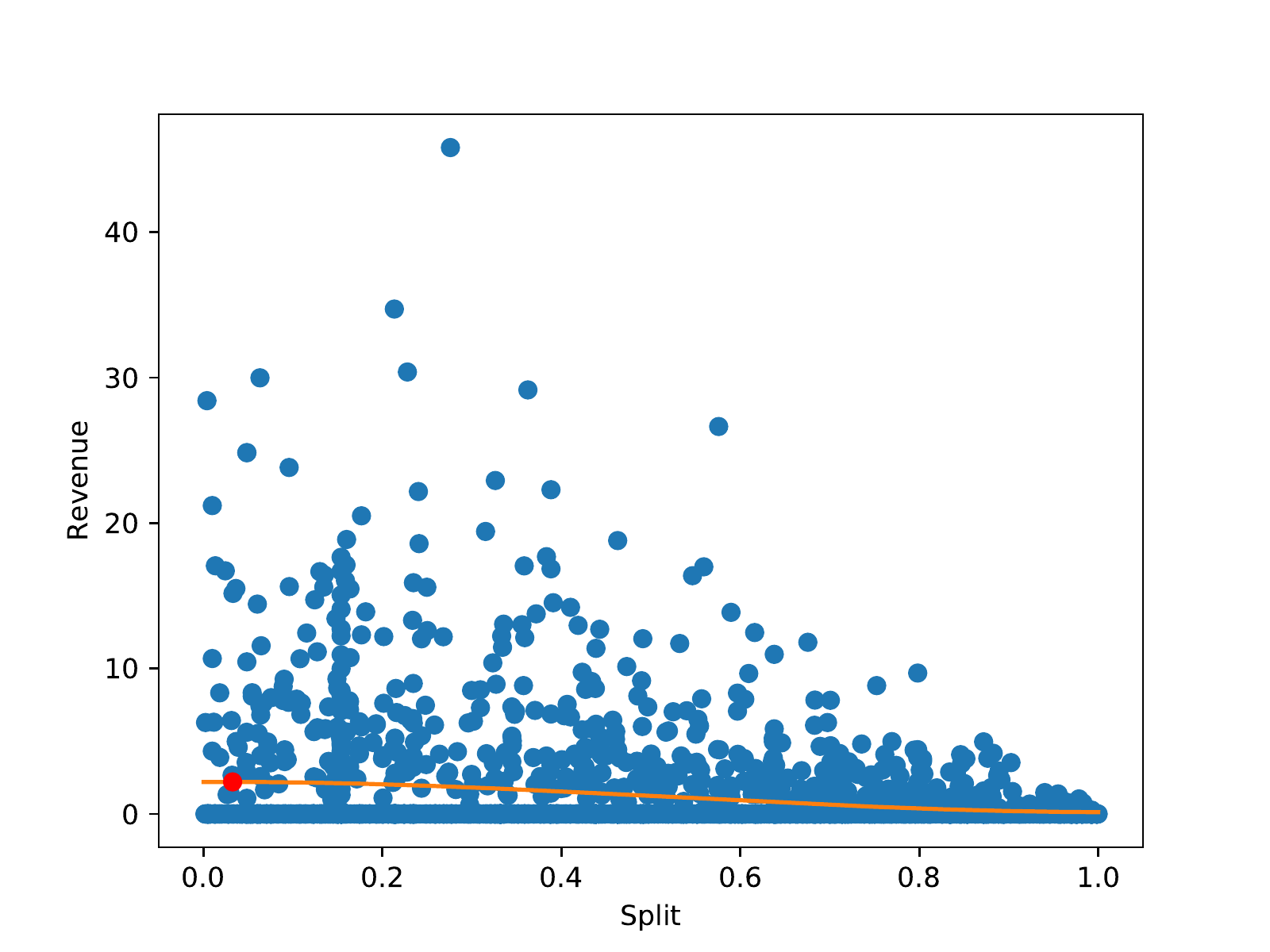}
\caption{Revenue of the participating rebate company as a function of the proposed split.}
\label{fig:bb}
\end{figure}

We collected a data stream $S$ in which the split proportion was selected randomly, $ a_t \sim \text{unif}(0,1)$, and used this data stream to evaluate different CAB policies. The offered discount to the customer was $y_t = 10 a_t $ and the reward of the CAB policies are some function of the proposed discount, $r = f(y)$. Again using \texttt{StreamingBandit} -- this time deployed in the field -- we obtained a data set consisting of a total of $2448$ data points (each consisting of a split $a_t$, and the actual revenue $r_t$). Figure \ref{fig:bb} shows the revenue of the rebate company against these random splits.\footnote{Note that the Figure seems to favor very low splits; this might not be feasible in the long run. Our current analysis only considers a single shot purchase, and obviously a more reasonable and sustainable approach would include the customer lifetime value as a whole; most likely favoring higher customer discounts.} 

\begin{figure}[htb]
\centering
\includegraphics[width=0.75\textwidth]{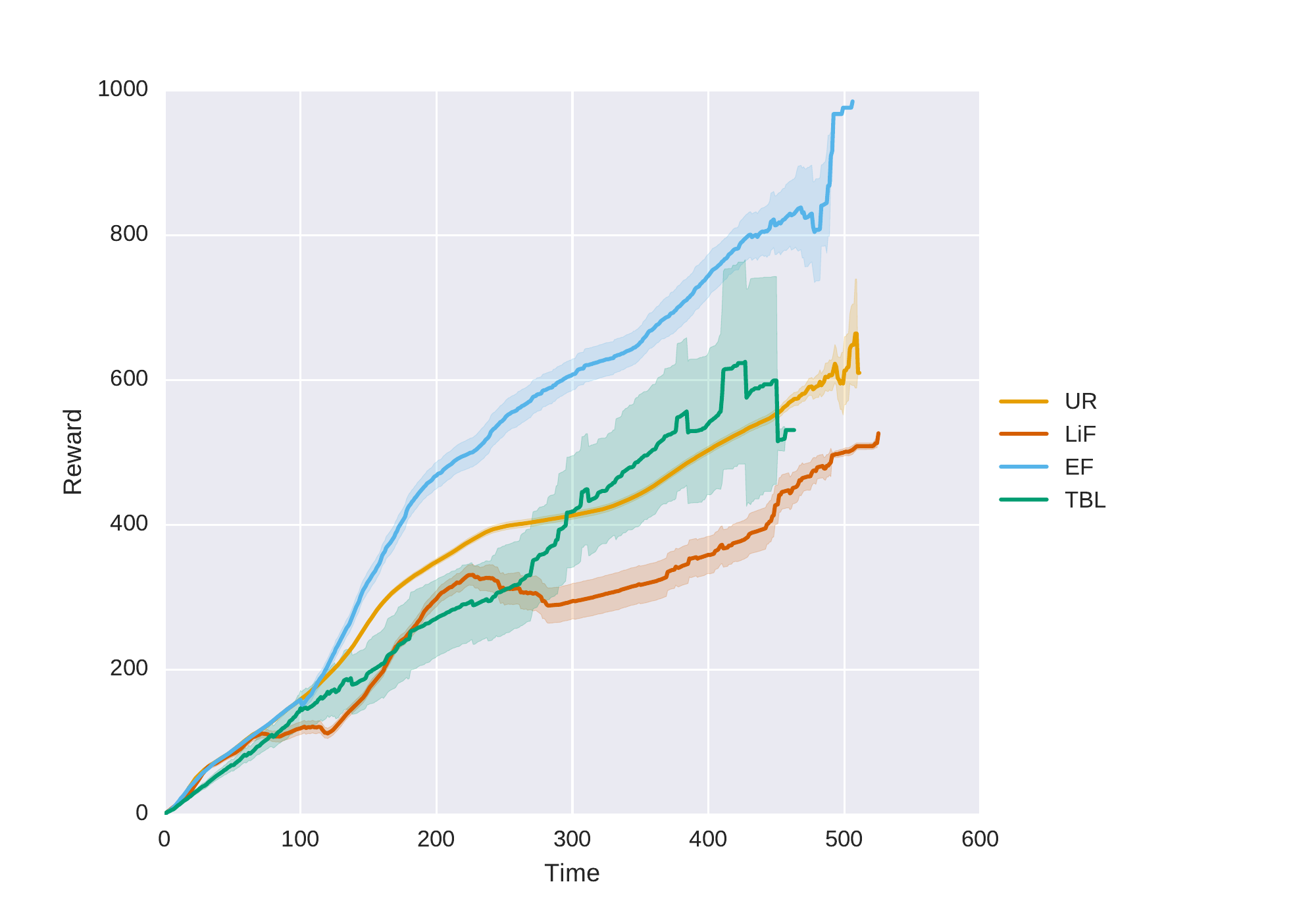}
\caption{Average cumulative reward and confidence bounds of four different CAB policies using our offline evaluation method.}
\label{fig:bbeval}
\end{figure}

As in our simulation studies, we run an offline evaluation using the empirical data $1,000$ times. We choose $\delta = 0.1$; this leads in expectation to around $500$ valid observations which aligns roughly with the median number of visits the rebate company expects per newly introduced product (often the rebate offers are valid only for specific products and for a limited period). We have used the same starting values for the policies as in the simulation studies, except for EF, where we limit the exploratory phase to $N = 100$. Note that since we have no knowledge about the action with the highest expected reward, we can only compute the cumulative reward, as discussed before. Figure \ref{fig:bbeval} shows the cumulative reward of the $4$ policies also used in our simulation study. The evaluations show that EF has obtains the highest cumulative reward. This analysis thus would encourage the company to use EF for the optimization of their revenue. This result is in-line with earlier studies that show that for small-scale problems simple heuristics often outperform asymptotically optimal policies \citep{kuleshov2014algorithms}. 

\section{Conclusion}

In this paper we proposed and empirically evaluated an offline policy evaluation method for the CAB problem that is inspired by the work by \citet{Li2011}. The method works by sequentially running through logged events while comparing the logged action with the action suggested by a policy under evaluation. Next, the distance between the suggested action and logged action is compared and, if this distance is small enough, the data point gets accepted and evaluated. We showed that using this method the rank ordering of CAB policies stays relatively intact and hence that the method is potentially of use to select CAB policies for applied problems. In future studies, incorporating more reasonable competitors for the current policies would increase the strength of this method further. Our work, however, raises a number of questions. 

Firstly, it seems that the ranking of policies is influenced by the complexity of the data generating model. Apparently, some policies are more robust to the noise introduced by our offline evaluation method than others and this interacts with our offline evaluation method; we will be looking for ways to quantify this difference and possibly correct for it. 

Secondly, we would like to scrutinize further the effect of the tuning parameter $\delta$, and perhaps quantify the behavior of distinct policies as $\delta$ decreases: as $\delta \rightarrow 0$ the actual (or online) behavior of the policy should surface and hence it is interesting to study the behavior of policies as a function of $\delta$. 

Furthermore, especially in relation to the introduced noise (first remark) and the scrutinization of $\delta$ (second remark), we did not consider a multi-dimensional action space, in which case the quality of the estimation of $E(R_c)$ might be even more dependent on the chosen delta and the smoothness of the expected reward. Future research should ideally also take this into account.  

Thirdly, there is currently no clear guidance on how to set $\delta$. We must stress, however, that with our method, one can explore multiple values of $\delta$ for the offline evaluation and re-run it, since the the choice of $\delta$ does not impact the collection of the training data. We have described the trade-off between a large and small $\delta$ before and we think it is useful to explore the robustness of the policies using multiple choices for $\delta$. Nonetheless, a clear cut way of choosing $\delta$ would make this method that more user friendly.

Finally, we would like to further study ways in which the noise introduced by the approximate evaluation of the true data generating model can be corrected. While currently the distance between the suggested and the logged action is not taken into account when updating the policy, we are experimenting with methods that update the parameters of a policy using a weight that is proportional to this distance (e.g., update the parameters of the policy using a discount that is dependent on $|a - \Pi(h_{t-1})|$ (as in Algorithm \ref{alg:cab_eval})).  

Additionally to the remarks of our current research, we have looked at another possibility for extending the method by \citet{Li2011} to use for the CAB problem. This would be to bin the continuous action space, such that we create an offline policy evaluator that is comparable to the method by \citet{Li2011}. Firstly, our method can be considered a quite dynamic -- and hence more continuous -- form of binning (i.e., a bin centered around the evaluated action). Secondly, this introduces also introduces a hyperparameter (similar to $\delta$), determining the amount of bins, which would be (more) highly application dependent. Hence, we think that the solution proposed in this paper is more appropriate.

We content that the method and its empirical analysis presented in this paper provide an initial step towards the development of valid and effective \emph{offline} policy evaluation method for CAB policies.


\bibliographystyle{spbasic}      
\bibliography{refs}   

\end{document}